\documentclass{article}
\usepackage{ijcai24}

\usepackage{times}
\usepackage{soul}
\usepackage{makecell}
\usepackage{amsfonts}
\usepackage{url}
\usepackage[hidelinks]{hyperref}
\usepackage[utf8]{inputenc}
\usepackage[small]{caption}
\usepackage{graphicx}
\usepackage{amsmath}
\usepackage{amsthm,amsmath}
\usepackage{booktabs}
\usepackage{enumerate}
\usepackage{algorithm}
\usepackage{algorithmic}
\usepackage[switch]{lineno}
\usepackage{color}
\usepackage{mathtools}
\usepackage{enumitem}
\usepackage{multirow}
\title{MCM: Multi-condition Motion Synthesis Framework}

\author{
Zeyu Ling$^{1*}$,
Bo Han$^{1*}$,
Yongkang Wong$^2$,
Han Lin$^1$,
Mohan Kangkanhalli$^2$\And
Weidong Geng$^{1\dagger}$\\
\affiliations
$^1$College of Computer Science and Technology, Zhejiang University\\
$^2$National University of Singapore\\
\emails
zeyuling@zju.edu.cn,
borishan815@zju.edu.cn,
yongkang.wong@nus.edu.sg,
h0h972351@gmail.com,
mohan@comp.nus.edu.sg,
gengwd@zju.edu.cn
}

\begin{document}

\maketitle
\begin{figure*}[ht]
    \centering
    \includegraphics[width=0.95\textwidth]{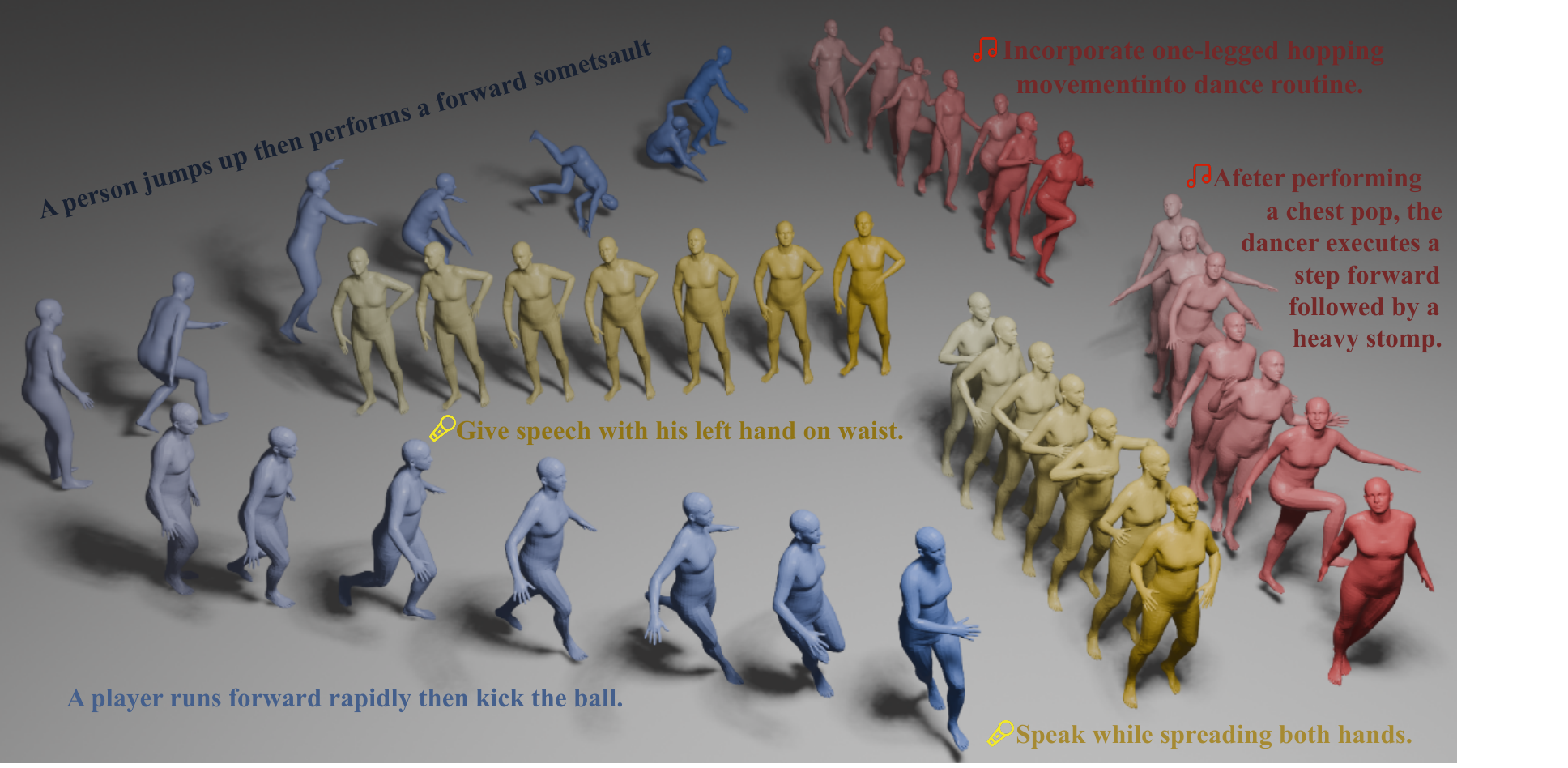}
    \caption{Samples of Multi-Condition Motion synthesis (MCM). MCM can generate human motion across various scenarios: text-to-motion (blue),  dance(red), and co-speech(yellow) motion synthesis under multiple conditions. The motion generated by MCM not only conforms to the rhythm of the music and speech but also exhibits consistency with the textual descriptions of dance and gestures.}
    \label{fig_main_vis}
\end{figure*}
\begin{abstract}
Conditional human motion synthesis (HMS) aims to generate human motion sequences that conform to specific conditions. Text and audio represent the two predominant modalities employed as HMS control conditions. While existing research has primarily focused on single conditions, the multi-condition human motion synthesis remains underexplored. In this study, we propose a multi-condition HMS framework, termed MCM, based on a dual-branch structure composed of a main branch and a control branch. This framework effectively extends the applicability of the diffusion model, which is initially predicated solely on textual conditions, to auditory conditions. This extension encompasses both music-to-dance and co-speech HMS while preserving the intrinsic quality of motion and the capabilities for semantic association inherent in the original model.
Furthermore, we propose the implementation of a Transformer-based diffusion model, designated as MWNet, as the main branch. This model adeptly apprehends the spatial intricacies and inter-joint correlations inherent in motion sequences, facilitated by the integration of multi-wise self-attention modules.
Extensive experiments show that our method achieves competitive results in single-condition and multi-condition HMS tasks.

\end{abstract}
\section{Introduction}
\footnote{* Equal contribution \textdagger Corresponding Author}
Human motion synthesis finds extensive applications in fields such as film production, game development, and simulation. Traditional manual animation techniques are notably constrained in terms of efficiency. 

Following the advent of neural network-based generative models, a variety of these models, such as Variational Autoencoders (VAEs)~\cite{kingma2013auto}, Generative Adversarial Networks~\cite{goodfellow2014generative}, Denoising Diffusion Probabilistic Models (DDPM)~\cite{ho2020denoising} have been adapted and refined for the specific domain of human motion generation to achieve high-fidelity results.

Current conditional human motion generation, such as text-to-motion~\cite{guo2022generating,zhang2022motiondiffuse} and music-to-dance~\cite{siyao2022bailando,tseng2023edge} focuses on generating human motion from a singular modality condition. 
The integration of condition information from different modalities is an area that has not been extensively explored. The dual-modality-driven 3D motion generation task presents the following primary challenges:
\begin{itemize}[leftmargin=3mm]
    \item Current HMS datasets predominantly encompass unimodal conditions. For instance, HumanML3D~\cite{guo2022generating} and AIST++~\cite{li2021ai} respectively include only text and music modalities. 
    \item Text and audio, being temporally non-aligned modalities, pose significant challenges in ensuring both temporal and semantic correlation with 3D human motion spontaneously.
    \item Fine-tuning directly from an existing single-condition pre-trained model may necessitate structural adjustments to the existing model architecture on one hand, and, on the other hand, fine-tuning may lead to catastrophic forgetting of the pre-trained conditional association capabilities.
\end{itemize}
Typically, TM2D~\cite{Gong_2023_ICCV} and UDE~\cite{Zhou_2023_CVPR} tokenize motion from HumanML3D~\cite{guo2022generating} and AIST++~\cite{li2021ai} using Motion VQ-VAE~\cite{van2017neural}. In the generation phase, they use text and sound respectively to drive GPT~\cite{hudson2021generative} to generate motion token sequences and then integrate these sequences by using weighted merging or replacement methods to achieve multi-condition driven. Employing the post-fusion approach circumvents the need to construct large-scale text-sound-motion data pairs; however, it harbors a critical drawback: each motion token in the sequence is essentially generated under the influence of a single modality condition. Tokens generated based on text do not conform to auditory conditions, while those generated from sound fail to align with semantic information.
	
To address the above challenges, we propose a novel end-to-end framework \textbf{MCM} (\textbf{M}ulti-\textbf{C}ondition \textbf{M}otion synthesis), which is tailored for multi-conditon driven 3D human motion synthesis.
MCM adopts a dual-branch structure, comprising the main branch and the control branch. Initially, the main branch leverages an arbitrary pre-trained text2motion DDPM network like MotionDiffuse~\cite{zhang2022motiondiffuse} and MDM~\cite{tevet2022human}, ensuring the motion quality and semantic coherence during multi-condition motion synthesis as shown in Figure \ref{fig_main_vis}. Subsequently, the control branch initializes its parameters, mirroring the structure of the main branch, and assumes the responsibility of modifying the motion in accordance with the auditory conditions. During the training of tasks conditioned on sound, MCM freezes the parameters of the main branch and activates those of the control branch. This approach is designed to preserve the motion quality and semantic correlation capabilities of the main branch without compromise. This methodology obviates the need for collecting a text-audio-motion dataset, yet enables both audio and text conditions to simultaneously influence every part of the motion sequence.

Regarding the structural design of the main branch, existing text2motion models~\cite{zhang2022motiondiffuse,tevet2022human} employ self-attention along the temporal dimension to grasp the sequential associations of motion sequences and perform cross-attention with textual descriptions to capture semantic information. However, when considering motion sequences, it is imperative to recognize that the channel dimension holds valuable spatial information and inter-joint relationships within the human body, aspects that have often been underappreciated. Therefore, we propose a Transformer-based DDPM network \textbf{MWNet} that incorporates a~\textbf{M}ulti-\textbf{W}ise self-attention mechanism (channel-wise self-attention) to better learn spatial information. 
	
In summary, our core contributions are as follows:
\begin{itemize}
    \item We introduce a unified MCM framework for 3D human motion synthesis based on multiple conditions. Remarkably, without necessitating structural reconfiguration of the network, MCM extends the capabilities of DDPM-based methods to sound-conditional inputs.
    \item We propose a Transformer-based architecture MWNet, enriched with a multi-wise attention mechanism, which leverages spatial information within motions to achieve better motion generation quality and comparable semantic matching capacity.
    \item Extensive experiments show that our method demonstrates competitive performance in the single-condition-driven tasks (text2motion and music2dance) and multi-condition-driven tasks. Furthermore, we also present an ablation analysis to elucidate the contribution of each component, enhancing understanding of their individual and collective impact on the system's performance.
\end{itemize}

\section{Related Work}
	
\subsection{Text-to-motion}

Text-to-motion converts descriptive text into motion sequences using various generative models. VAE-based methods~\cite{guo2022generating,guo2022tm2t,petrovich2022temos} learn motion distribution in latent space but are limited by VAE's Gaussian posterior estimation, resulting in suboptimal results. Recently, diffusion models have shown impressive image synthesis capabilities, prompting increased research into their application for motion generation.

MotionDiffuse~\cite{zhang2022motiondiffuse} and MDM~\cite{tevet2022human} employ DDPM for human motion synthesis. MotionDiffuse features a linear self-attention mechanism, whereas MDM uses a vanilla Transformer Encoder and predicts the original motion sequence at each time step, instead of step-specific noise. MLD~\cite{chen2023executing} utilizes a latent vector DDPM for forward noising and reverse denoising in motion latent space. MAA~\cite{azadi2023make} enhances performance on out-of-distribution data by pre-training a diffusion model on a large dataset of (text, static pseudo-pose) pairs. ReMoDiffuse~\cite{Zhang_2023_ICCV} introduces a retrieval-enhanced DDPM, significantly improving in-distribution text-to-motion capabilities.

With advancements in Large Language Models, autoregressive generative models are now used to link two different modalities. TM2T~\cite{guo2022tm2t}, T2MGPT~\cite{zhang2023t2m}, and MotionGPT~\cite{jiang2023motiongpt} approach text-to-motion by using VQ-VAE~\cite{van2017neural} to discretize motion sequences and employ GRU~\cite{chung2014empirical}, GPT~\cite{vaswani2017attention,radford2019language}, or T5~\cite{2020t5} models to learn the correlations between motion and text tokens.

\subsection{Music-to-dance}

The music-to-dance task involves generating dance movements synchronized with music beats. DanceNet~\cite{zhuang2022music2dance} uses an LSTM-based auto-regressive method to create dance movements. DeepDance~\cite{sun2020deepdance} employs Generative Adversarial Networks for human motion generation. FACT~\cite{li2021ai} introduces the Full-Attention Cross-modal Transformer that utilizes self-attention among music, dance movements, and combined music-dance tokens. Bailando~\cite{siyao2022bailando} utilizes a VQ-VAE encoding for separate upper and lower body parts, integrating them into full-body dance sequences using GPT. EDGE~\cite{tseng2023edge} employs a Transformer-based DDPM for precise music-rhythm matching and introduces a contact consistency loss to address physical errors in foot movements. EnchantDance~\cite{han2023enchantdance} develops a robust dance latent space to train a dance diffusion model effectively.

\subsection{Multi-condition HMS}
The aforementioned method has a limitation in that it can only accept a single modality condition as input, which is insufficient to meet the requirements of many scenarios. Some methods have made efforts to integrate multiple conditions and accommodate various scenarios. TriModal~\cite{yoon2020speech} and CaMN~\cite{liu2022beat} generate co-speech motion based on speech and spoken text, with CaMN integrating additional conditions such as facial expressions. MoFusion~\cite{dabral2023mofusion} for the first time, implements a unified UNet-based DDPM architecture, which, through two different types of conditional encoders, can effectively function in both text-to-motion and music-to-dance tasks. UDE~\cite{zhou2023ude} and TM2D~\cite{gong2023tm2d} both adopt a model structure based on VQ-VAE and GPT and propose a similar approach to integrate music and text conditions. They generate motion token sequences conditioned on sound and text respectively, and then either weight or concatenate the generated logits to obtain a smooth motion sequence that blends motion tokens conditioned on both text and audio.

\section{Method}
\subsection{Problem definition} 
\label{sec:problem}
The objective of the multi-condition human motion synthesis task is to generate a motion sequence $X \in \mathbb{R}^{T \times D}$ under a set of constraint conditions $C$. $X$ is an array of $x_{i}$, where $i \in \{1, 2, \ldots, T\}$, and $T$ denotes the number of frames. Each $x_{i} \in \mathbb{R}^{D}$ represents the D-dimensional pose state vector at the $i$-th frame. $c_{j} \in C$  could be textual description, speech voice, or background music.

\begin{figure*}[t]
    \centering
    \includegraphics[width=0.95\textwidth]{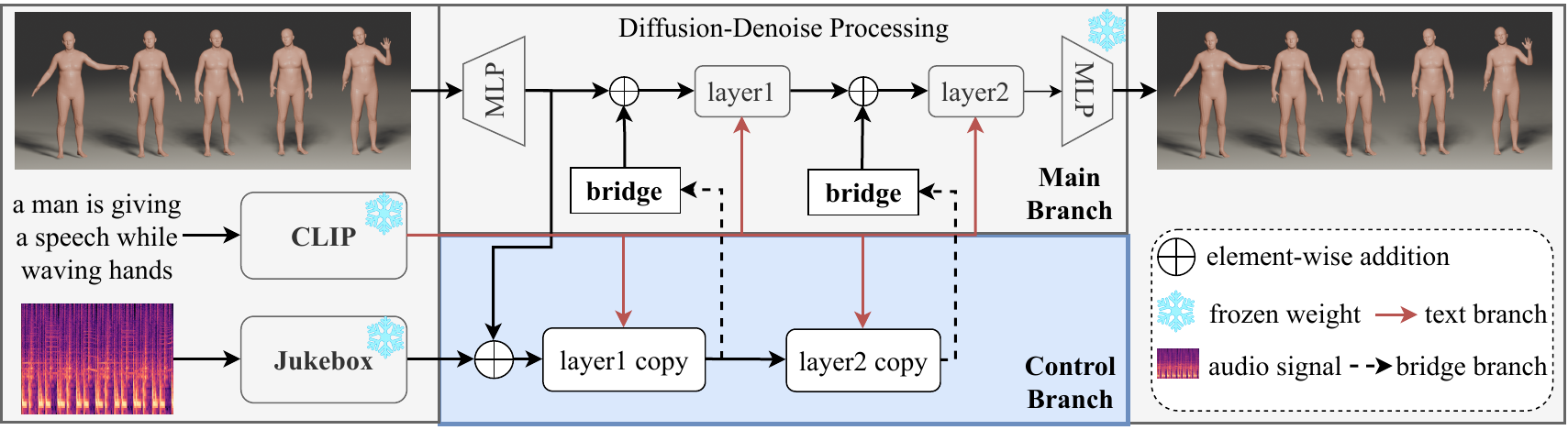}
    \caption{Overview of the simplified two-layer MCM framework. MCM employs a dual-branch structure consisting of the main branch and the control branch. The layer-wise outputs from the control branch are connected to the main branch via bridge modules, which are fully connected layers or 1d-convolutions with parameters initialized to zero. The output of each bridge module is directly added to the input feature vector of corresponding layers in the main branch. The condition encoders encompass several pre-trained feature extractors for different modal conditions.}
    \label{fig_overall_arch}
\end{figure*}

\subsection{MCM Framework} 
As depicted in Figure~\ref{fig_overall_arch}, the MCM framework employs a dual-branch architecture, comprising a pre-trained main branch with frozen parameters and a trainable control branch. The main branch is composed of any neural network grounded in DDPM, such as MotionDiffuse or MDM. In this study, we have devised and implemented a DDPM network named MCM as the main branch of MCM, which will be comprehensively detailed in Section \ref{sec:mwnet}. The control branch mirrors the structural framework of the main branch and is initialized utilizing the parameters derived from the main branch.

Drawing inspiration from ControlNet~\cite{zhang2023adding}, we implement a strategy of optimizing the main branch and control branch independently during distinct stages of the training process. 
We pre-train the main branch on the text-to-motion task, with the objective of this phase being to endow the MCM with foundational motion quality and semantic association capabilities. In the training phase of audio-to-motion tasks, all parameters, excluding those assigned to the control branch and bridge module, are maintained in a fixed state. This strategy is implemented to guarantee the retention of the main branch's generative quality and its capabilities in semantic association. 

For the input to the Control branch, we perform element-wise addition of the Jukebox features with the motion latent vector to incorporate audio information into MCM. 
The output of each layer of the Control branch is added to the input of the main branch through the bridge module, thereby causing a slight offset in the output of the main branch under the control of audio conditions. By performing zero-initialization on the bridge module, we ensure that the initial output of MCM is identical to that of the main branch and gradually adjust the parameters according to the audio conditions during the iterative process.
\subsection{MWNet Architecture}
\label{sec:mwnet}
As previously elucidated, the quality of the main branch exerts a substantial impact on the final quality and semantic association capacities of the motions generated by MCM. Consequently, we delved deeper into the structural aspects of the main branch in the context of the text-to-motion task and proposed the MWNet model.

\begin{figure}[t]
    \centering
    \includegraphics[width=0.9\columnwidth]{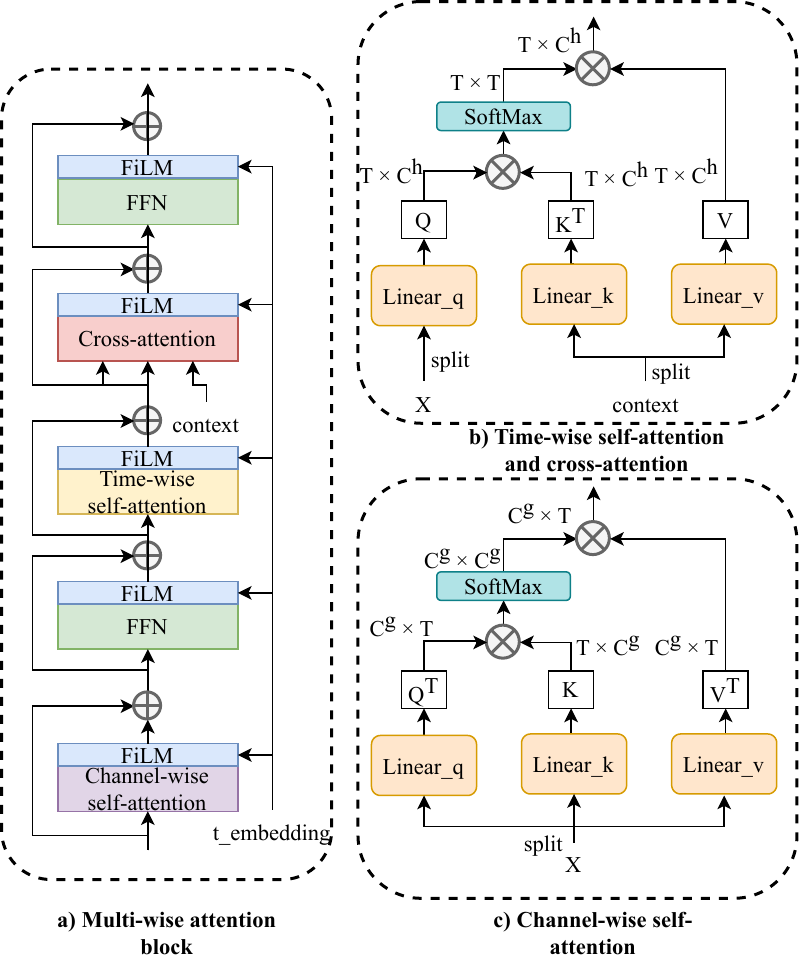} 
    \caption{Model architecture for a multi-wise attention block. It incorporates three distinct types of attention modules, which are employed alternately. The symbols ``+'' and ``×'' separately represent feature addition and multiplication operations. $T$ symbolizes the length of the input sequence, while $C^g$ and $C^h$ signify the number of channels for the matrices $Q$, $K$, and $V$ after. The split operation means splitting the channels into $g$ groups or $h$ heads. Context represents text condition for cross-attention and is exactly equal to $X$ for time-wise self-attention.} 
    \label{fig_arch}
\end{figure}

As delineated in Section~\ref{sec:problem}, the channel dimension of the motion sequence encompasses the entirety of the spatial information pertinent to the motion sequence. Nevertheless, current DDPM-based motion generation models, including MotionDiffuse and MDM, predominantly concentrate on time-wise self-attention and cross-attention mechanisms. These are employed to model the correlations at the temporal level and the semantic level between motions and their respective conditions. In the spatial dimension, information is currently modeled exclusively through linear layers and activation functions. We posit that this approach is inadequate for capturing the complex spatial relationships present in intricate motion sequences, such as the correlations between joints. Consequently, we advocate for the use of channel-wise self-attention~\cite{ding2022davit} and introduce the Multi-Wise attention mechanism, specifically designed to model these critical aspects of spatial information. MWNet is composed of layers arranged in the configuration depicted in Figure \ref{fig_arch} (a).

Similar to Stable Diffusion~\cite{rombach2022high} and GLIDE~\cite{nichol2021glide}, we use FiLM~\cite{perez2018film} blocks to furnish timestamp information to MWNet after every attention or Feed Forward Network (FFN) modules. In the FiLM module, the weights and biases for the affine transformation performed on the motion latent vector are derived from the timestep mapping, as shown in the following:
\begin{equation}
    FiLM(x, \epsilon_t)=x+LN(x\odot(W_1+I)\epsilon_t)+W_2\epsilon_t
\end{equation}
$LN$ denotes layer normalization module~\cite{ba2016layer}. $W_1$, and $W_2$ are two projection matrices. $I$ denotes a matrix wherein all elements are 1, possessing a shape congruent with that of $x$. $\odot$ denotes the element-wise multiplication.

With projection weights $W^Q$, $W^K$, $W^V$, $X$ are projected to $Q=XW^Q$, $K=XW^K$, $V=XW^V$ and split into $N_h$ heads or $N_g$ groups. We denote $Q_i$, $K_i$, and $V_i$ for each head or group. Time-wise self-attention can be denoted as follows:
\begin{equation}
    SA_T(Q_i,K_i,V_i)=Softmax(\frac{Q_iK_i^T}{\sqrt{C_h}})V_i
\end{equation}
\begin{equation}
    SA_T(Q,K,V)=\{SA_T(Q_i,K_i,V_i)\}_{i=0}^{N_h}
\end{equation}
Whereas, channel-wise self-attention can be denoted as:
\begin{equation}
    SA_C(Q_i,K_i,V_i)=(Softmax(\frac{Q_i^TK_i}{\sqrt{C_g}})V_i^T)^T
\end{equation}
\begin{equation}
    SA_C(Q,K,V)=\{SA_C(Q_i,K_i,V_i)\}_{i=0}^{N_g}
\end{equation}

Where $C_h$ and $C_g$ denote the channel numbers for each head or group.

\section{Experiments}
\subsection{Dataset} 
To train on various datasets, we process the motion of all involved datasets (HumanML3D, AIST++, and BEAT) to the same format with a 22-joint skeleton (the first 22 joints of the SMPL skeleton) and 20 FPS. Subsequently, following HumanML3D, we use a 263-dimension representation $x=concat(\dot{r}^a,\dot{r}^x,\dot{r}^z,r^y,j^p,j^v,j^r,c^f)$ to represents motions at every frame. $\dot{r}^a \in \mathbb{R}$ is root angular velocity along the Y-axis; $\dot{r}^x,\dot{r}^z \in \mathbb{R}$ are root linear velocities on XZ-plane; $r^y$ is root height; $j^p \, j^v \in \mathbb{R}^{3j}$ and $j^r \in \mathbb{R}^{6j}$ are the local joints positions, velocities, and rotations in root space, with $j$ denoting the number of joints; $c^f \in \mathbb{R}^4$ is binary features obtained by thresholding the heel and toe joint velocities to emphasize the foot ground contacts. 

We also processed the AIST++ dataset~\cite{li2021ai} and the BEAT dataset~\cite{liu2022beat} to generate pseudo text descriptions, thereby forming audio-text-motion sample pairs for training in the control stage. 

\subsection{Evaluation metrics}
For the text2motion evaluation, we employ previously established metrics~\cite{guo2022generating} to assess the generated motion sequences across various dimensions. (1) Motion quality: Frechet Inception Distance (FID)~\cite{onuma2008fmdistance} evaluates the dissimilarity between two distributions by calculating the difference between feature vector (extracted by feature extractor) statistical measures (mean and covariance). (2) Diversity: The Diversity and MultiModality metrics respectively assess the degree of variation in the generated motions with different text inputs and with the same text input. (3) Semantic matching: The motion-retrieval precision (R Precision) evaluates the accuracy of matching between texts and motions using Top 1/2/3 retrieval accuracy. Multi-modal Distance (MM Dist) measures the distance between motions and text features extracted by the feature extractor trained with contrastive learning. 

For the music2dance, we adopt the commonly used evaluation metrics following~\cite{dabral2023mofusion}. (1) FID: utilizing kinetic and geometry features implemented within fairmotion~\cite{gopinath2020fairmotion}. (2) Diverisy: it computes the average pairwise Euclidean distance of the kinetic and geometry features of the motions generated from music in the validation and test set. (3) Beat Alignment Score (BAS): a metric that quantifies the congruence between kinematic beats and musical beats. Kinematic beats correspond to the local minima of kinetic velocity within a motion sequence, signifying points where motion momentarily halts. 

\subsection{Implementation Details} 
We conduct training of MCMs utilizing distinct DDPM-like main branch architecture, including MotionDiffuse~\cite{zhang2022motiondiffuse}, MDM~\cite{tevet2022human}, and our MWNet. The conditioning inputs from diverse modalities are pre-processed through the employment of pre-trained condition encoders. We use CLIP~\cite{radford2021learning} to extract features from text conditions. Similar to EDGE, we use the prior layer of Jukebox~\cite{dhariwal2020jukebox} to extract features from all audio conditions(music and human voice) and downsample them to the same frame rate as the motion samples (20 FPS).  Regarding the diffusion model, we set the number of diffusion steps at 1000, while the variances $\beta_t$ follow a linear progression from 0.0001 to 0.02. We employ the Adam optimizer for training the model, employing a learning rate of 0.0002 throughout both training phases. In concordance with MDM, we adopt the strategy of predicting $x_{start}$ as an alternative to the prediction of noise. This approach is aimed at achieving an enhanced quality of motion. 

\subsection{Text-to-Motion Evaluation} 
We train and evaluate our main branch model MWNet on HumanML3D~\cite{guo2022generating} dataset.
\begin{table*}[t]
    \centering
    \small
    \begin{tabular}{lccccccc}
        \toprule
        \multirow{2}{*}{\centering Methods} & \multicolumn{3}{c}{R Precision $\uparrow$} & \multirow{2}{*}{\centering FID $\downarrow$} & \multirow{2}{*}{\centering MultiModal Dist $\downarrow$} & \multirow{2}{*}{\centering Diversity $\rightarrow$} & \multirow{2}{*}{\centering MultiModality $\uparrow$} \\
        \cmidrule(r){2-4}
        &Top 1 & Top 2 & Top 3 & & & & \\
        \midrule
        Real motions & 0.511$^{\pm.003}$ & 0.703$^{\pm.003}$ & 0.797$^{\pm.002}$ & 0.002$^{\pm.000}$ & 2.974$^{\pm.008}$ & 9.503$^{\pm.065}$ & - \\
        \midrule
        
        T2M & 0.457$^{\pm.002}$ & 0.639$^{\pm.003}$ & 0.740$^{\pm.003}$ & 1.067$^{\pm.002}$ & 3.340$^{\pm.008}$ & 9.188$^{\pm.002}$ & 2.090$^{\pm.083}$ \\

        TM2T & 0.424$^{\pm.003}$ & 0.618$^{\pm.003}$ & 0.729$^{\pm.002}$ & 1.501$^{\pm.017}$ & 3.467$^{\pm.011}$ & 8.589$^{\pm.076}$ & 2.424$^{\pm.093}$\\
        
        TEMOS & 0.424$^{\pm.002}$ & 
        0.612$^{\pm.002}$ & 0.722$^{\pm.002}$ & 3.734$^{\pm.028}$ & 3.703$^{\pm.008}$ & 8.973$^{\pm.071}$ & 0.368$^{\pm.018}$ \\
        
        TM2D & - & - & 0.319 & 1.021 & 4.098 & \textcolor{red}{9.513} & \textcolor{red}{4.139} \\
        
        T2MGPT & 0.491$^{\pm.003}$ & 0.680$^{\pm.003}$ & 0.775$^{\pm.002}$ & 0.116$^{\pm.004}$ & 3.118$^{\pm.011}$ & 9.761$^{\pm.081}$ & 1.856$^{\pm.011}$ \\

        MotionGPT & 0.492$^{\pm.003}$ & 0.681$^{\pm.003}$ & 0.778$^{\pm.002}$ & 0.232$^{\pm.008}$ & 3.096$^{\pm.008}$ & \textcolor{blue}{9.528}$^{\pm.071}$ & 2.008$^{\pm.084}$ \\

        AttT2M & 0.499$^{\pm .003}$ & 0.690$^{\pm .002}$ & 0.786$^{\pm .002}$ & 0.112$^{\pm .006}$ & 3.038$^{\pm .007}$ & 9.700$^{\pm .090}$ & 2.452$^{\pm .051}$ \\
        
        \midrule

        MoFusion & - & - & 0.492 & - & - & 8.82 & 2.521 \\
        
        MLD & 0.481$^{\pm.003}$ & 0.673$^{\pm.003}$ & 0.772$^{\pm.002}$ & 0.473$^{\pm.013}$ & 3.196$^{\pm.010}$ & 9.724$^{\pm.082}$ & 2.413$^{\pm.079}$ \\

        MDM & 0.320$^{\pm .005}$ & 0.498$^{\pm 0.004}$ - & 0.611$^{\pm .007}$ & 0.544$^{\pm .044}$ & 5.566$^{\pm .027}$ & 9.559$^{\pm .086}$ & \textcolor{blue}{2.799}$^{\pm .072}$ \\
        
        MotionDiffuse & 0.491$^{\pm.001}$ & 0.681$^{\pm.001}$ & 0.782$^{\pm.001}$ & 0.630$^{\pm.001}$ & 3.113$^{\pm.001}$ & 9.410$^{\pm.049}$ & 1.553$^{\pm.042}$ \\

        ReMoDiffuse & \textcolor{red}{0.510\(^{\pm.005}\)} & \textcolor{red}{0.698\(^{\pm.006}\)} & \textcolor{red}{0.795\(^{\pm.004}\)} & \textcolor{blue}{0.103}\(^{\pm.004}\) & \textcolor{red}{2.974\(^{\pm.016}\)} & 9.018\(^{\pm.075}\) & 1.795\(^{\pm.043}\) \\

        \textbf{MWNet(ours)} & \textcolor{blue}{0.502}$^{\pm.002}$ & \textcolor{blue}{0.692}$^{\pm.004}$ & \textcolor{blue}{0.788}$^{\pm.006}$ & \textcolor{red}{0.053}$^{\pm.007}$ & \textcolor{blue}{3.037}$^{\pm.003}$ & 9.585$^{\pm.082}$ & 0.8104$^{\pm.023}$ \\
        
        \bottomrule
    \end{tabular}
    \caption{Quantitative results on the HumanML3D test set. The symbol $\rightarrow$ denotes that the results are more favorable when the metric closely approximates the distribution of real motions (i.e., the metrics of authentic movements). The methodologies are categorized based on their reliance on DDPM. A demarcation line is utilized to distinguish the approaches, with those situated below the line being DDPM-based. Red font indicates the best results, while blue denotes the second. Due to the inherent randomness of the metrics, most methods were evaluated twenty times to calculate the mean and variance (as superscripts), while TM2D and MoFusion didn't provide the variance.} 
    \label{table_eval_text2motion}
\end{table*}

Table \ref{table_eval_text2motion} presents the quantitative metrics of our method on the HumanML3D dataset. We achieved the best motion quality and the second semantic relevance among all methods. Moreover, it surpasses all non-DDPM methods. Owing to its retrieval-based mechanism, ReMoDiffuse remains the model with the most robust semantic correlation capabilities. Compared to the TM2D and MoFusion approach, which similarly accommodates the integration of auditory conditions, our method exhibits a distinct advantage in both semantic relevance and the quality of motion. A robust capability for semantic association lays the groundwork for maintaining strong semantic relational abilities in multi-condition HMS scenarios. Our method exhibits moderate performance on the MultiModality metric. The underlying causes of this phenomenon were explored in our ablation studies.

\subsection{Music-to-Dance Evaluation} 
During the training phase on the AIST++ dataset, we processed the training data into motion segments with a maximum length of 196 frames, equivalent to 9.8 seconds. In the evaluation stage, prior studies were assessed using the complete dance motion sequences from the validation and test sets of AIST++, maintaining the original frame rate of 60 FPS. To facilitate an equitable comparison with these studies, we linearly interpolated the generated motion sequences, initially at 20 FPS, to upscale them to 60 FPS. Table \ref{table_eval_music2dance} presents the evaluation outcomes of MCM on the AIST++ dataset. When utilizing MCM, three distinct DDPM methodologies, previously trained on the text2motion dataset, all demonstrated performance on par with state-of-the-art (SOTA) methods in the music2dance task. As demonstrated in the table ~\ref{table_eval_music2dance}, our approach achieved the best $FID_k$, while UDE achieved the best $FID_g$, and the diversity of the dance movements generated by each method was found to be remarkably similar. Furthermore, our methodology, while preserving a comparable level of dance motion quality and diversity, attains SOTA metrics in beat alignment exclusively via the elementary element-wise integration of Jukebox music features. Among various implementations utilizing the MCM framework, MWNet + MCM demonstrates relatively superior dance motion quality and beat alignment. Conversely, MotionDiffuse + MCM excels in achieving enhanced diversity in its outputs.

\begin{table}
    \centering
    \tiny
    \begin{tabular}{lcccccc}
        \toprule
        Methods & $\mathrm{FID_k}\downarrow$ & $\mathrm{FID_g}\downarrow$ & $\mathrm{Div_k}\rightarrow$ &$\mathrm{Div_g}\rightarrow$ & BAS$\uparrow$ \\
        \midrule
        
        Ground Truth & 17.10 & 10.60 & 8.19 & 7.45 & 0.2374 \\
        
        \midrule
        DanceNet & 69.18 & 25.49 &2.86 & 2.85 & 0.143 \\
        
        DanceRevolution & 73.42 & 25.92 & 2.86 & 3.52 & 0.195 \\

        FACT & 35.35 & 22.11 & 5.94 & 6.18 & 0.2209 \\

        Bailando & 28.16 & \textcolor{blue}{9.62} & \textcolor{red}{7.83} & 6.34 & 0.233 \\

        UDE & \textcolor{blue}{17.25} & \textcolor{red}{8.69} & \textcolor{blue}{7.78} & 5.81 & 0.231 \\

        TM2D & 19.01 & 20.09 & 9.45 & \textcolor{blue}{6.36} & 0.204 \\

        MoFusion & 50.31 & - & 9.09 & - & 0.253 \\
        
        EDGE & - & 23.04 & - & - & 0.270 \\
        
        \midrule
        
        MotionDiffuse + finetune & 45.22 & 13.21 & 7.27 & 5.39 & \textcolor{blue}{0.273} \\
        
        MotionDiffuse + MCM  & 44.27 & 13.43 & 9.24 & 5.29 & 0.266 \\

        MDM + finetune & 47.39 & 22.07 & 8.94 & 5.16 & 0.258 \\
        
        MDM + MCM & 39.21 & 19.55 & 5.81 & 6.34 & 0.265 \\
        
        MWNet + finetune & 34.73 & 20.25 & 5.87 & \textcolor{red}{6.79} & 0.249 \\

        MWNet + MCM & \textcolor{red}{15.57} & 25.85 & 6.50 & 5.74 & \textcolor{red}{0.275} \\
        
        \bottomrule
    \end{tabular}
    \caption{Results on AIST++ validation and test set. The method labeled "finetune" employs the single-branch structure introduced in section ~\ref{sec:ablation}, replacing the dual-branch structure of MCM.}
    \label{table_eval_music2dance}
\end{table}

\subsection{Text-sound multi-condition Generation}

\begin{figure}
    \centering
    \includegraphics[width=0.9\columnwidth]{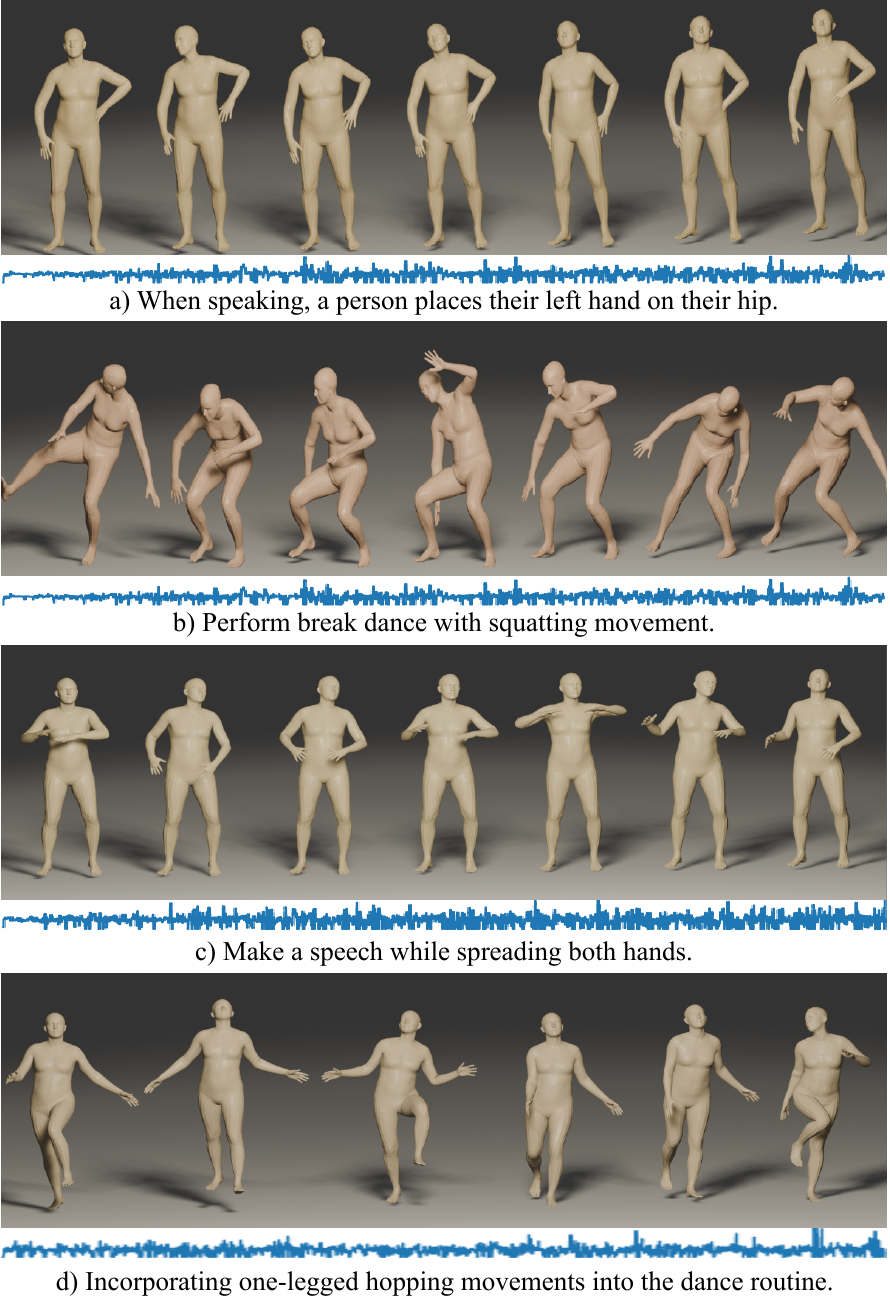}
    \caption{Text-sound multi-condition motion synthesis with MCM (MWNet as the main branch). Each sample is obtained using a segment of text and a segment of audio as inputs.}
    \label{fig:multi_condition}
\end{figure}

As illustrated in Figure \ref{fig:multi_condition}, within the multi-condition HMS, we generate corresponding dance or co-speech movements under the joint control of textual and auditory conditions. For instance, we adjust specific gestures during a speech and particular dance moves in a dance through text, achieving interesting control effects. 

\begin{table*}[tb]
    \centering
    \small
    \begin{tabular}{lccccccc}
        \toprule
        \multirow{2}{*}{\centering Methods} & \multicolumn{3}{c}{R Precision $\uparrow$} & \multirow{2}{*}{\centering FID$\downarrow$}  & \multirow{2}{*}{\centering MultiModal Dist$\downarrow$} & \multirow{2}{*}{Diversity $\rightarrow$} & \multirow{2}{*}{\centering MultiModality $\uparrow$} \\
        \cmidrule(r){2-4}
        & Top 1 & Top 2 & Top 3 & & & & \\
        \midrule
        T/CA/F & 0.389 & 0.547 & 0.668 & 1.081 & 3.772 & \textbf{9.393} & \textbf{3.033} \\
        T/CA/F/T/F & 0.423 & 0.600 & 0.708 & 0.825 & 3.577 & 9.018 & 2.788 \\
        T/F/T/CA/F & 0.431 & 0.617 & 0.720 & 0.586 & 3.472 & 9.157 & 2.152 \\
        T/CA/F/CS/F & \textbf{0.455} & \textbf{0.642} & \textbf{0.744} & 0.751 & 3.399 & 8.933 & 1.707 \\
        CS/F/T/CA/F(MWNet) & \textbf{0.455} & 0.640 & 0.742 & \textbf{0.377} & \textbf{3.349} & 9.312 & 1.481 \\
        \bottomrule
    \end{tabular}
    \caption{The results on the HumanML3D dataset after training Transformer Decoder modules with different structures for 500 epochs.}
    \label{mwnet_arch_compare}
\end{table*}
In order to conduct a fair comparison with existing methods capable of integrating textual and auditory conditions. We sampled 10 music segments from the AIST++ dataset and 5 segments from outside the AIST++ dataset, and designed distinct dance motion descriptions for each piece of music. We employed TM2D and MCM to generate dance movements under multi-conditional constraints, specifically, to produce dance sequences that conform to textual descriptions and align with musical rhythms. Subsequently, we invited 20 users to conduct a comparative assessment of the generated outcomes. The subjects were asked to make comparisons from three aspects: beat alignment, text match, and motion quality. Importantly, the entire evaluation process was meticulously designed to be anonymous, thereby ensuring that participants were unable to discern the origins of the data samples, which in turn guaranteed an unbiased assessment of the different models. The results of the user study reveal a distinct preference among users for our outcomes on the AIST++ dataset. Moreover, our results also demonstrate a relative advantage in external datasets. This is attributed to the fact that in MCM, both text and audio conditions concurrently influence the entire motion sequence. In contrast, TM2D's post-fusion approach results in each motion token being driven effectively by only a single modality. Consequently, the actions generated from the text do not align with the rhythm of the music. Simultaneously, our superiority in semantic relevance is largely attributable to the significantly stronger semantic correlation capabilities of our main branch compared to TM2D, as evident from the evaluation results on the HumanML3D dataset.

\begin{table}[bt]
    \centering
    \small
    \begin{tabular}{cccc}
    \toprule
    \multirow{2}{*}{\centering Data} & \multicolumn{3}{c}{User Study}  \\ 
    \cmidrule(r){2-4}
     & beat align & text match & motion quality \\
    \midrule
     
    AIST++ & 95.0\% & 95.5\% & 93.0\% \\ 
                            
    Wild & 69.0\% & 74.0\% & 62.0\% \\  
                             
    \bottomrule
    \end{tabular}
    \caption{We request users to compare the quality of the multi-condition HMS from three aspects: beat alignment, dance motion quality, and semantic match. Each cell in the table represents the win rate of MCM in each comparative aspect.}
    \label{tab:multi_condition}
\end{table}

\subsection{Ablation Study}
\label{sec:ablation}
\subsubsection{Dual-branch vs single-branch}
In this section, we further explore the superiority of the Dual-branch structure in transferring from text-condition scenarios to sound-condition scenarios. The dual branch structure trains the main branch on the text2motion task, and the control branch on the sound condition; whereas for the single branch structure, we directly optimize the parameters of the main branch during the control stage, and the audio condition is added directly to the motion latent vector as input. In the lower half of Table~\ref{table_eval_music2dance}, we compare the performance differences on the AIST++ dataset between dual-branch and single-branch structures equipped with different main branches. The results show that all dual-branch outcomes surpass their single-branch counterparts. Additionally, the single-branch structure no longer possesses the semantic association obtained from text-to-motion pretraining.

\subsubsection{Multi-wise self-attention module design}
We designed experiments to verify the necessity and superiority of multi-wise self-attention module. Our Multi-wise self-attention module consists of a time-wise self-attention (denoted as TA), a channel-wise self-attention (CS), a cross-attention (CA), and two FFNs (F). By modifying or reordering these modules, we compared the performance metrics of different configurations on the HumanML3D benchmark. 
In this experiment, each trial was limited to training for only 500 epochs to conserve experimental time. We experimented with arranging or adjusting the attention modules for different categories. As shown in Table \ref{mwnet_arch_compare}, comparing T/CA/F/T/F and T/CA/F/CS/F, it can be observed that substituting CS for T results in significantly improved motion quality, as CS compensates for the spatial information deficiencies of T. It was found that using both T and CS in the module simultaneously helps significantly in enhancing semantic relevance and motion quality. Additionally, we observed that placing CS before T and CA can improve motion quality with a slight loss in semantic relevance. We also found that the downside of CS is that it leads to reduced MultiModality, which we aim to explore and address in future work. Ultimately, we chose the option that offered the highest motion quality as the module design of MWNet.

\section{Conclusion}
We propose MCM, a novel framework for the multi-conditioned motion synthesis method that spans multiple scenarios. Utilizing MCM, methods based on DDPM for text2motion tasks can concurrently adapt to multi-conditions without necessitating any structural modifications. Additionally, we introduce a Transformer-based architecture MWNet that incorporates multi-wise self-attention, enhancing the modeling of spatial information and inter-joint correlations. We quantitatively evaluate our approach across tasks based on various modal conditions. In the text-to-motion task, our results surpass all existing methods in terms of motion quality and are comparable in semantic correlation capabilities. In the music-to-dance task, we achieved the best beat alignment and $FID_k$ metrics, along with comparable dance diversity and quality. In the multi-condition HMS domain, our approach accomplishes the concurrent generation of dance movements based on both text and sound, as well as co-speech motion.

\section*{Acknowledgments}
This work was supported in part by the Zhejiang Province (China) Research Project 2023C01045 and National Natural Scientific Foundation of China 61972346.
\bibliographystyle{named}
\bibliography{ijcai24}

\end{document}